\documentclass[conference]{IEEEtran}
\IEEEoverridecommandlockouts
\usepackage{amsmath,amssymb,amsfonts}
\usepackage{algorithmic}
\usepackage{graphicx}
\usepackage{hyperref}
\usepackage{textcomp}
\usepackage{xcolor}
\usepackage[
backend=biber,
style=ieee,
]{biblatex}
\def\BibTeX{{\rm B\kern-.05em{\sc i\kern-.025em b}\kern-.08em
    T\kern-.1667em\lower.7ex\hbox{E}\kern-.125emX}}
\begin{document}

\title{Survey of Machine Learning Techniques To Predict Heartbeat Arrhythmias}

\author{\IEEEauthorblockN{Samuel Armstrong}
\IEEEauthorblockA{\textit{Institute for Biomedical Informatics} \\
\textit{University of Kentucky Healthcare}\\
Lexington, Kentucky, USA \\
Sam.Armstrong@uky.edu}
}
\maketitle

\begin{abstract}
Many works in biomedical computer science research use machine learning techniques to give accurate results. However, these techniques may not be feasible for real-time analysis of data pulled from live hospital feeds. In this project, different machine learning techniques are compared from various sources to find one that provides not only high accuracy but also low latency and memory overhead to be used in real-world health care systems.
\end{abstract}

\begin{IEEEkeywords}
electrocardiogram, prediction, AI, timeseries classification
\end{IEEEkeywords}

\section{Introduction}
Many medical research institutions are producing massive amounts of data. So much, that researchers, clinicians, and lab technicians are overwhelmed when identifying and classifying data points in real-time. This can lead to fatigue or errors that ultimately could be avoided with the help of an AI-assisted application; specifically, a system to find and learn trends in data that would not necessarily be found otherwise. Using machine learning can help all aspects of data analysis and classification regarding healthcare workloads. The goal of this project is to explore machine learning techniques on electrocardiogram (ECG) datasets to reduce the workload on cardiologists and increase accuracy when determining heartbeat abnormalities.

\subsection{Datasets}
Two data sources were used in this project. The first is the Massachusetts Institute for Technology-Beth Israel Hospital (MIT-BIH) Arrhythmia database \cite{MIT-BIH, Physionet_std_citation}. This database is very well-established and includes over 100,000 unique, labeled heartbeats. Additionally, this set does not have a predefined split, therefore, a split of 20\% for testing, 20\% for validation, and 60\% for training was used. This set is used in articles \cite{art1} and \cite{art2}. 

The second is the Physikalisch-Technische Bundesanstalt (PTB) Diagnostic ECG database \cite{PTB, Physionet_std_citation} used in articles \cite{art2} and \cite{art3} and contains 549 12-lead records. This database also does not have a predefined split, and similarly to the MIT-BIH database, a split of 20\% for testing, 20\% for validation, and 60\% for training was used. 

\section{Related Works}
From studying related works in this field, convolutional neural networks (CNNs) (sometimes accompanied by a long-short term memory (LSTM) network) are one of the best candidates to classify ECG data \cite{art1,art2,art3}. These types of neural networks are much larger and inherently slower when compared to the k-Nearest Neighbors algorithm, for example. As explained in \cite{art3}, CNNs do not necessarily have to be highly complex to lift features from this type of data, especially when a time constraint is present. When handling only a single ECG lead the data becomes 1-dimensional, which can be processed and inferred upon with two to seven convolutional layers. 

However, in ECG recordings that include 12 leads, the data now becomes 2-dimensional. Mentioned by the authors in article  \cite{art3}, including all 12 leads (if available) increases the detection of bundle branch block and ventricular hypertrophy. By including most (if not all) 12 leads when training, the model has the ability to learn more finely ingrained features. 

Generally, the accuracy of a machine learning technique relies on the data used to train it. For example, a CNN model trained on a one dataset and tested on another is likely, if not guaranteed, to have a lower prediction performance. However, article \cite{art2} claims to use a method of model transfer to train on the MIT-BIH database and infer on the PTB database. This is similar to the environment that the model will operate in, so this method will be used.

\section{Problem Formation}
At the University of Kentucky Albert B. Chandler Hospital, an electronic intensive care unit (eICU) feed is being established, which will handle thousands of data points for each patient. These points range from oxygen saturation to pH in the blood. A preliminary sample of this data has been collected, which contains a 12-lead ECG at about one second resolution for each patient. The pipeline is being actively developed, so an increase in resolution should occur to nearly match that of the MIT-BIH and PTB databases. With this in mind and nearly 300 ICU patients, the amount of data per day will be overwhelming. To assist in processing and diagnoses, a machine learning method would be greatly beneficial to clinicians. Specifically, interest in using CNNs to accomplish this task has been expressed by multiple parties both academically and practically.

At the beginning of this project, the goal was to compare new machine learning techniques related to predicting heartbeat arrhythmias to decide which performed the best. This has since evolved into the creation of an model based on articles \cite{art1, art2, art3}. During the initial phase of this project, the only goal was to recreate the methods described in each of the three articles. This has proven to be an unfortunate challenge because two of the three are not open-source projects. After trying to recreate each of the article's methods, the performance metrics claimed in articles \cite{art1, art3} could not be achieved. The likely cause of this was not using the same methods to preprocess the data or set up the neural network during the recreation. However, there are takeaways from these articles and elements from each are present in the proposed model. From this point forward, it is clear that the method described in article \cite{art2} is a solid base.

Currently, a modified model based on \cite{art2} has been created with the top priorities being inference speed and accuracy. While the eICU data for ECG leads is still not available, the two databases (MIT-BIH and PTB) presented in this article will continue to be used. These databases are similar enough to live eICU data that using them should give an approximation of performance. These metrics are recorded in the Results section below. 

\section{Proposed Approach}
\subsection{Base Implementation}
\begin{figure}[h!]
  \centerline{\includegraphics[width=0.5\textwidth]{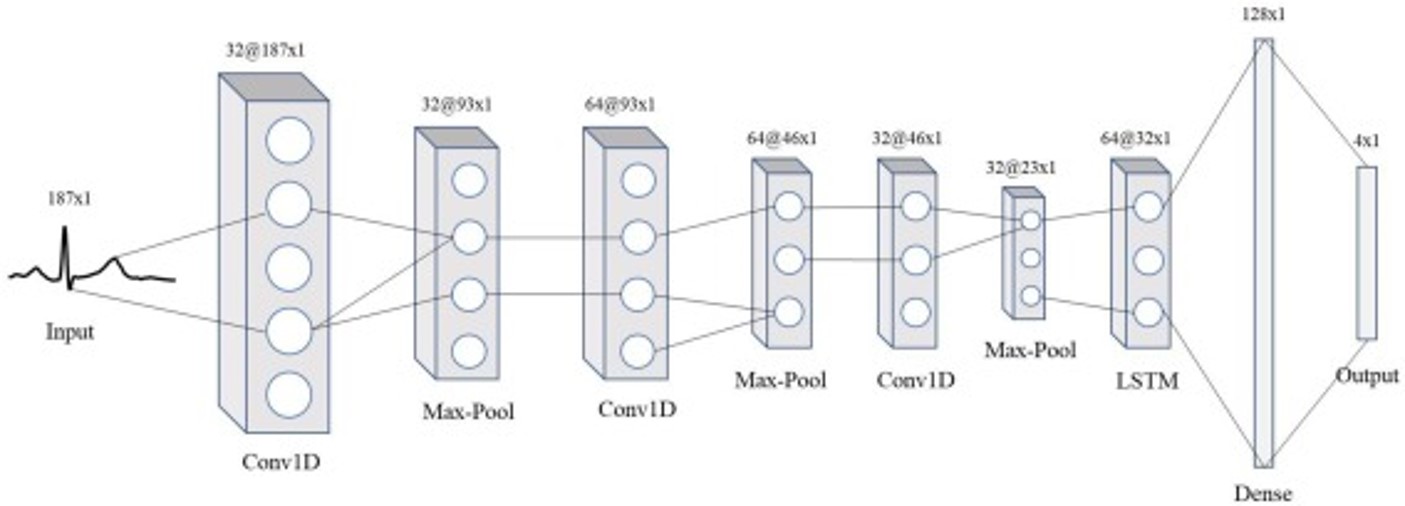}}
  \caption{Article \cite{art1} Proposed Model}
  \label{fig4}
\end{figure}

\begin{figure}[h!]
  \centerline{\includegraphics[width=0.5\textwidth]{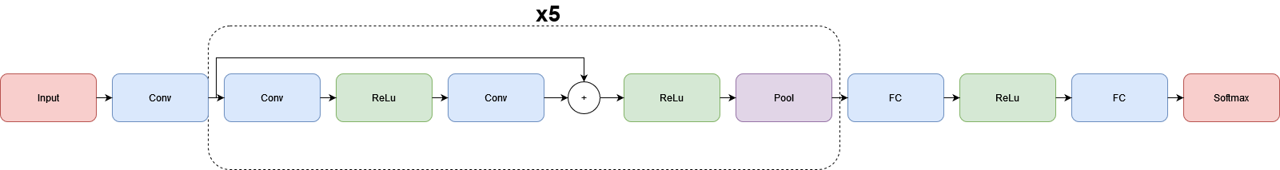}}
  \caption{Article \cite{art2} Proposed Model}
  \label{fig5}
\end{figure}

\begin{figure}[h!]
  \centerline{\includegraphics[width=0.5\textwidth]{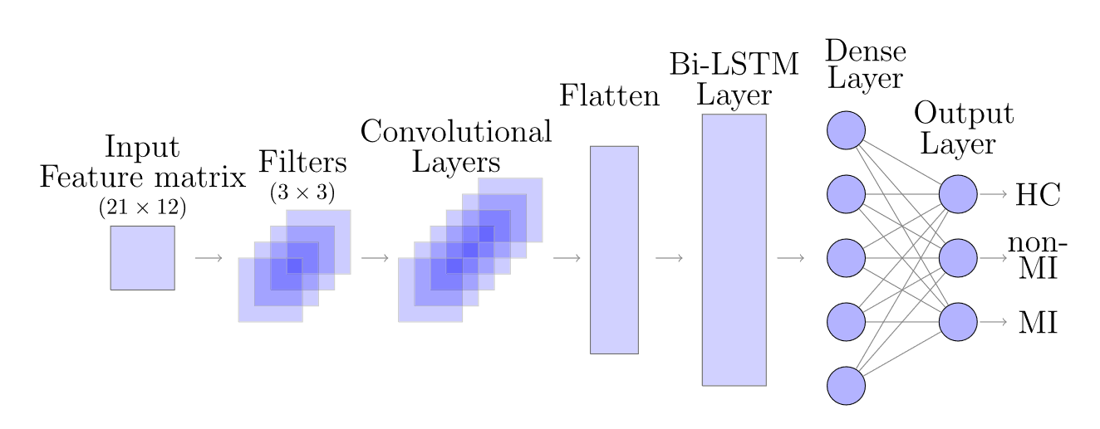}}
  \caption{Article \cite{art3} Proposed Model}
  \label{fig6}
\end{figure}

When this project began, the methods listed in articles \cite{art1, art2, art3} and shown in Figures \ref{fig4}, \ref{fig5}, and \ref{fig6} were going to be implemented, then compared to see which method was best for ECG inference. Comparison of these algorithms was based on these metrics: accuracy, F1 score, specificity, sensitivity, training time, time to infer, and ease of reproduction. While working through each of the three articles, it was clear that some aspects of the reproduction of the model would be a "best guess" implementation. Articles \cite{art1} and \cite{art3} did not go into enough detail to accurately capture the results that were claimed. For example, there are disparities in graphics showing the layering of the CNN when compared to the text describing the model. This made it difficult to accurately represent those article methods. As seen in the Results section, these two methods performed significantly worse than what is described in their respective articles. However, article \cite{art2} not only explains every detail of their preprocessing and model design, but also provides code to run the model on both the MIT-BIH and PTB databases. This allowed the easy reproduction of this model and verification of the results. Since this article produces a model that ranks highly in the proposed metrics, this method of training, testing, and model transfer was used for the remainder of the project. 

\subsection{Removal of Complexity}
After choosing the second article as a baseline for the remainder of this project, modifications to decrease training time and increase inference rate were implemented ensuring that the integrity of the metrics remained the same. After experimentation, there are three areas of this model that could be improved, the first of which is the complexity. The method described in article \cite{art2} uses 11 1-dimensional convolutional layers to build the model with a kernel size of 3. While the model has been verified to work well in this instance, the training time on the MIT-BIH database is 1033 seconds and testing achieved 6186 heartbeats/second. This testing rate is acceptable, but the training time is too long, especially if retraining occurs at a regular interval. The anticipated result of removing the five repeated hidden layers from the network (shown in Figure \ref{fig2}) is a decrease in training time while maintaining the metrics seen before modification. This may produce a substantially quicker time, because, as explained in article \cite{art1}, having only three 1-dimensional convolution layers produced a training time of 340 seconds. While this article did not produce the desired results, it shows that there is a proportional relationship between complexity, time, and accuracy. However, leaving six convolutional layers in the model should not substantially affect the metrics of the model. 

\subsection{Adaptation Plan Based on Modification Results}
If the metrics of the modified model underperform, then it is worthwhile to revisit the article's disregard of all ECG leads except for lead II. As explained in article \cite{art3}, the QRS section of a heartbeat is 0.08 to 0.11 seconds long. If a QRS is longer than this, it is an indication of an arrhythmia. However, a proper diagnosis of this type of arrhythmia can only be determined if a longer QRS is found in another lead. The use of most (if not all) leads could be beneficial to aid the feature extraction of a machine learning algorithm. Depending on the training time of the 12-lead data, a decision to return to a single lead or to individually add convolutional layers back to the model to find a satisfactory middle ground.

\section{Results}
The results included in this section are produced on a workstation PC with an Nvidia RTX A5000 GPU, an Intel i7-11700 CPU, and 16 GB RAM. 

It is important to note that both initial and modified (final) models were given an "EarlyStopping" callback provided by the Python package Keras. This callback stops training if a metric, such as validation loss or validation accuracy, plateaus. With this in mind, each of the following models stopped early in the range of 15 to 30 epochs. Therefore, based on the settings for this callback, early stopping may not halt on the same epoch every time. This will affect training time and inference rate. The settings used for this project are described in Table \ref{table:earlystop}.

\begin{table}[htbp]
\begin{center}
\caption{EarlyStopping Settings}
\begin{tabular}{|c|c|c|}
\hline
Article & \textbf{Metric} & \textbf{Patience} \\
\hline
\cite{art1} & Validation Loss & 8 \\
\hline
\cite{art2} & Validation Accuracy & 5 \\
\hline
\cite{art3} & Validation Loss & 8 \\
\hline
\end{tabular}
\label{table:earlystop}
\end{center}
\end{table}

Another Python Keras callback was used called "ReduceLROnPlateau" in each of the three articles. This callback reduces the learning rate of the model if a plateau in either validation loss or validation accuracy occurs. Again, the settings for this callback are described in Table \ref{table:reduceLR}.

\begin{table}[htbp]
\begin{center}
\caption{ReduceLROnPlateau Settings}
\begin{tabular}{|c|c|c|c|c|}
\hline
Article & \textbf{Metric} & \textbf{Patience} & \textbf{Factor} & \textbf{Mode} \\
\hline
\cite{art1} & Validation Loss & 3 & 0.1 & "auto" \\
\hline
\cite{art2} & Validation Accuracy & 3 & 0.1 & "max" \\
\hline
\cite{art3} & Validation Loss & 3 & 0.1 & "auto" \\
\hline
\end{tabular}
\label{table:reduceLR}
\end{center}
\end{table}

\subsection{Initial Results (mid-project)}
As stated in the Problem Formation, there were issues of reproducibility of articles \cite{art1} and \cite{art3}. Therefore, their metrics are much lower than the proposed metrics in their respective papers. Tables \ref{table:genresults}, \ref{table:detart1results}, \ref{table:detart2results}, and \ref{table:detart3results} contain the results produced when recreating the respective article methods. 

\begin{table}[htbp]
\begin{center}
\caption{General Results for \cite{art1,art2,art3} Models}
\begin{tabular}{|c|c|c|c|}
\hline
Article & \textbf{\textit{Accuracy}} & \textbf{\textit{Training Time (sec)}}& \textit{\textbf{\begin{tabular}[c]{@{}l@{}}Inference Rate \\ (samples/sec)\end{tabular}}} \\
\hline
\cite{art1} & 0.80 & 653 & 4960 \\
\hline
\cite{art2} & 0.98 & \multicolumn{1}{c|}{\begin{tabular}[c]{@{}c@{}}1033 + 64 \\ (MIT-BIH to PTB)\end{tabular}}&  6186 \\
\hline
\cite{art3} & 0.65 & 573 & 872 \\
\hline
\end{tabular}
\label{table:genresults}
\end{center}
\end{table}

\begin{table}[htbp]
\begin{center}
\caption{Article \cite{art1} Detailed Results}
\begin{tabular}{|c|c|c|c|c|}
\hline
Class & \textbf{\textit{Precision}} & \textbf{\textit{Recall}}& \textbf{\textit{F1-Score}}& \textbf{\textit{Support}} \\
\hline
Normal & 0.95 & 0.83 & 0.88 & 44238 \\
\hline
Atrial Fibrillation & 0.23 & 0.08 & 0.12 & 1836 \\
\hline
Atrial Flutter & 0.79 & 0.89 & 0.84 & 3221 \\
\hline
AV Junctional Rhythm & 0.00 & 0.07 & 0.01 & 388 \\
\hline
\end{tabular}
\label{table:detart1results}
\end{center}
\end{table}

\begin{table}[htbp]
\begin{center}
\caption{Article \cite{art2} Detailed Results}
\begin{tabular}{|c|c|c|c|c|}
\hline
\multicolumn{5}{|c|}{MIT-BIH Only} \\
\hline
Class & \textbf{\textit{Precision}} & \textbf{\textit{Recall}}& \textbf{\textit{F1-Score}}& \textbf{\textit{Support}} \\
\hline
Normal & 0.99 & 1.00 & 0.99 & 18118 \\
\hline
Supraventricular & 0.92 & 0.75 & 0.83 & 556 \\
\hline
Ventricular & 0.97 & 0.94 & 0.96 & 1448 \\
\hline
Fusion Beat & 0.87 & 0.75 & 0.80 & 162 \\
\hline
Unknown & 0.99 & 0.98 & 0.99 & 1608 \\
\hline
\multicolumn{5}{|c|}{PTB Only} \\
\hline
Normal & 0.98 & 0.98 & 0.98 & 809 \\
\hline
Abnormal & 0.99 & 0.79 & 0.89 & 2102 \\
\hline
\multicolumn{5}{|c|}{MIT-BIH Model Transfer to PTB} \\
\hline
Normal & 0.98 & 0.99 & 0.98 & 809 \\
\hline
Abnormal & 0.99 & 0.99 & 0.99 & 2102 \\
\hline
\end{tabular}
\label{table:detart2results}
\end{center}
\end{table}

\begin{table}[htbp]
\begin{center}
\caption{Article \cite{art3} Detailed Results}
\begin{tabular}{|c|c|c|c|c|}
\hline
Class & \textbf{\textit{Precision}} & \textbf{\textit{Recall}}& \textbf{\textit{F1-Score}}& \textbf{\textit{Support}} \\
\hline
Healthy Control & 0.71 & 0.89 & 0.79 & 964 \\ 
\hline
 Non-Myocardial Infarction & 0.65 & 0.40 & 0.50 & 547 \\ 
\hline
Myocardial Infarction & 0.52 & 0.50 & 0.51 & 652 \\ 
\hline
\end{tabular}
\label{table:detart3results}
\end{center}
\end{table}

\subsection{Final Results}
After initially removing the hidden convolutional layers as described in the Proposed Approach section, there was a significant drop in performance from around 98\% accuracy to about 89\% on the MIT-BIH data. This was caused by a jump from 32 to 256 filters which originally increased in increments of base 2 (16, 32, 64, 128, 256). After removing the hidden layers, the new increments were 16, 32, 256. This issue was remedied by reducing the number of filters on the last two convolutions to 64. This proved to increase the accuracy back up to around 96-99\% after running five additional times. As seen in Tables \ref{table:genmodresults}, \ref{table:detmodresultsmit}, \ref{table:detmodresultsptb}, and  \ref{table:detmodresultsmitptb}, all metrics are nearly the same, if not better, as they were previously, especially regarding inference rates. 

\begin{figure}[h!]
  \centerline{\includegraphics[width=0.5\textwidth]{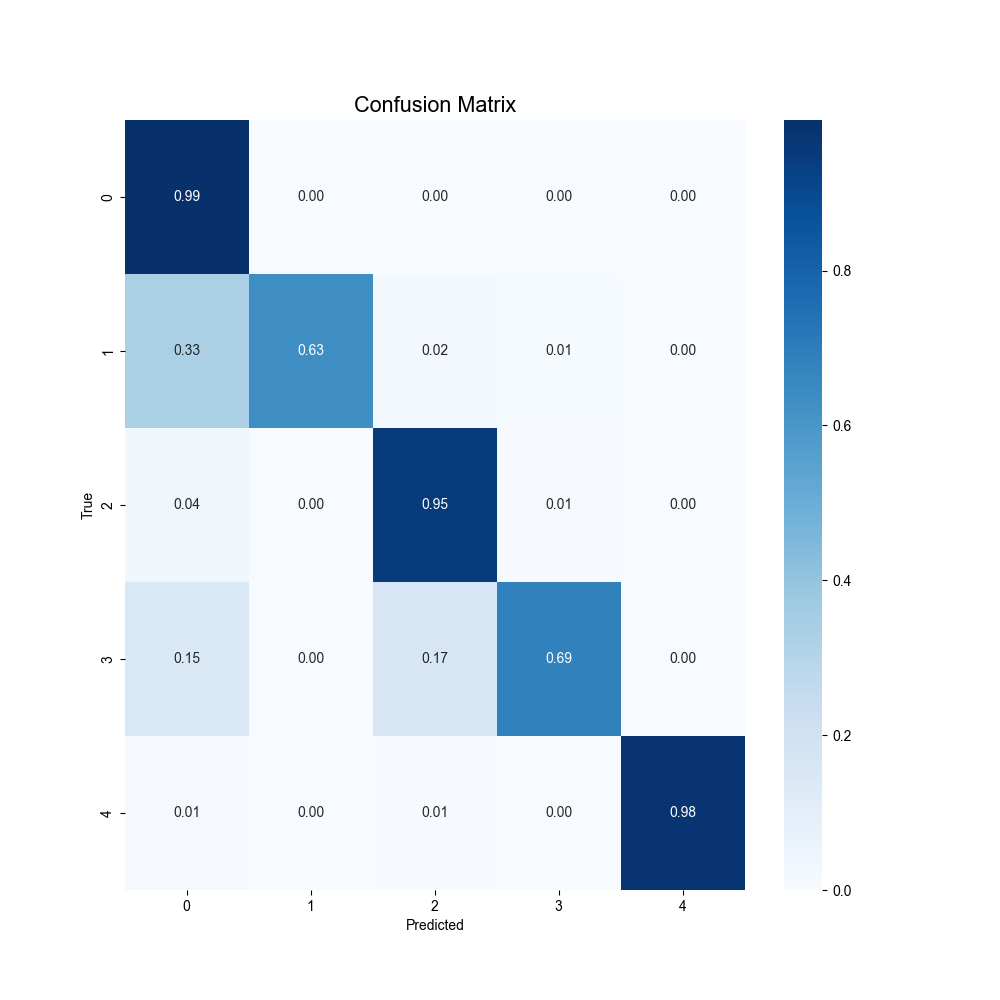}}
  \caption{Modified MIT-BIH Model Confusion Matrix}
  \label{fig1}
\end{figure}

\begin{figure}[h!]
  \centerline{\includegraphics[width=0.5\textwidth]{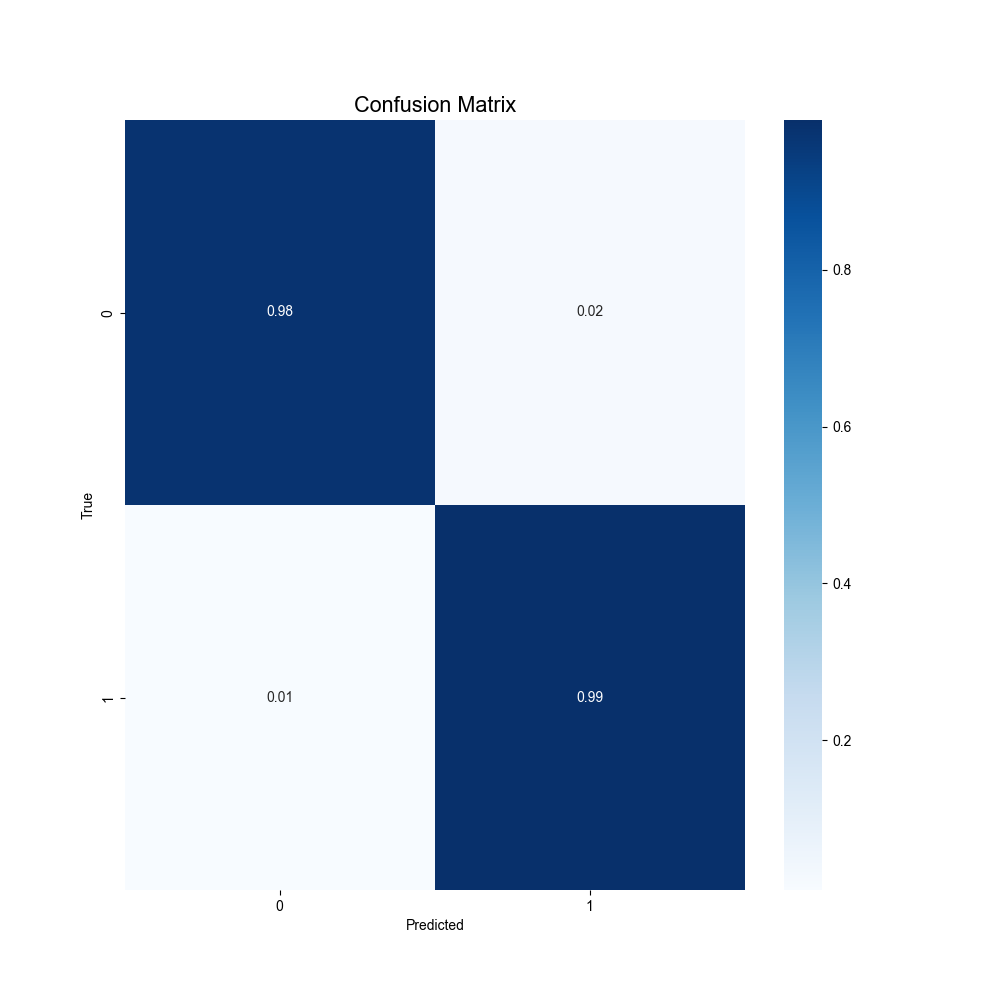}}
  \caption{Modified PTB Model Confusion Matrix}
  \label{fig2}
\end{figure}

\begin{figure}[h!]
  \centerline{\includegraphics[width=0.5\textwidth]{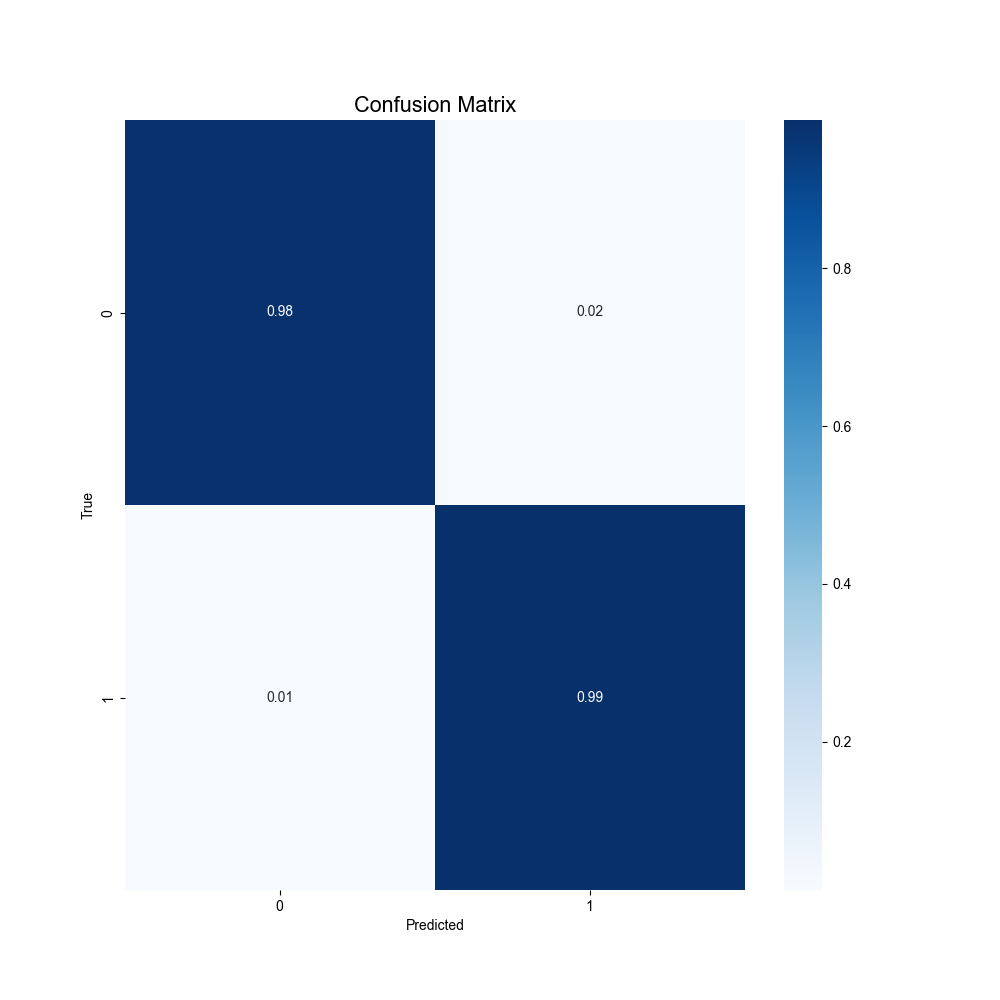}}
  \caption{Modified MIT-BIH to PTB Model Transfer Confusion Matrix}
  \label{fig3}
\end{figure}

Figures \ref{fig1}, \ref{fig2}, and \ref{fig3} show the confusion matrix of the modified models, which defines the inference performance of a classification mode. In Figure \ref{fig1}, the modification used to speed up the MIT-BIH model did produce some ambiguity in classes one and three. This is possibly attributed to the relatively low number of examples for training for these classes. To rectify this in the future, a 3-class classifier may be used instead of five classes as described in article \cite{art2}. However, depending on the workload a 5-class classifier may be needed, in which case, adding the hidden layers back into the original article method should produce better results. In Figures \ref{fig2} and \ref{fig3}, the performance when classifying two classes is exceptional and matches the metrics of the unmodified article method. Depending on the workload, solely using the PTB model or MIT-BIH to PTB transfer model for binary classification may not fit the scope. The MIT-BIH database offers more flexibility in that area.

\begin{table}[htbp]
\begin{center}
\caption{General Results for Modified Models}
\begin{tabular}{|c|c|c|c|}
\hline
Dataset & \textbf{\textit{Accuracy}} & \textbf{\textit{Training Time (sec)}}& \textit{\textbf{\begin{tabular}[c]{@{}l@{}}Inference Rate \\ (samples/sec)\end{tabular}}} \\
\hline
MIT-BIH & 0.98 & 728 & 10807 \\
\hline
PTB & 0.99 & 99 & 7374 \\
\hline
MIT-BIH, PTB & 0.99 & \multicolumn{1}{c|}{\begin{tabular}[c]{@{}c@{}}357 + 76 \\ (MIT-BIH to PTB)\end{tabular}} & 8274 \\
\hline
\end{tabular}
\label{table:genmodresults}
\end{center}
\end{table}

\begin{table}[htbp]
\begin{center}
\caption{Modified Model Results using MIT-BIH Database Only}
\begin{tabular}{|c|c|c|c|c|}
\hline
Class & \textbf{\textit{Precision}} & \textbf{\textit{Recall}}& \textbf{\textit{F1-Score}}& \textbf{\textit{Support}} \\
\hline
0 & 0.98 & 0.99 & 0.99 & 18118 \\
\hline
1 & 0.88 & 0.63 & 0.74 & 556 \\
\hline
2 & 0.94 & 0.95 & 0.94 & 1448 \\
\hline
3 & 0.85 & 0.69 & 0.76 & 162 \\
\hline
4 & 0.99 & 0.98 & 0.98 & 1608 \\
\hline
\end{tabular}
\label{table:detmodresultsmit}
\end{center}
\end{table}

\begin{table}[htbp]
\begin{center}
\caption{Modified Model Results using PTB Database Only}
\begin{tabular}{|c|c|c|c|c|}
\hline
Class & \textbf{\textit{Precision}} & \textbf{\textit{Recall}}& \textbf{\textit{F1-Score}}& \textbf{\textit{Support}} \\
\hline
0 & 0.98 & 0.98 & 0.98 & 809 \\
\hline
1 & 0.99 & 0.99 & 0.99 & 2102 \\
\hline
\end{tabular}
\label{table:detmodresultsptb}
\end{center}
\end{table}

\begin{table}[htbp]
\begin{center}
\caption{Modified Model Transfer Results using MIT-BIH and PTB Databases}
\begin{tabular}{|c|c|c|c|c|}
\hline
Class & \textbf{\textit{Precision}} & \textbf{\textit{Recall}}& \textbf{\textit{F1-Score}}& \textbf{\textit{Support}} \\
\hline
0 & 0.97 & 0.98 & 0.98 & 809 \\
\hline
1 & 0.99 & 0.99 & 0.99 & 2102 \\
\hline
\end{tabular}
\label{table:detmodresultsmitptb}
\end{center}
\end{table}

\section{Conclusion}
Based on the above results, it is reasonable to conclude that the method described in article \cite{art2} is best for practical use. Reducing the number of hidden layers and adjusting filter size produces a faster model that can be trained and used to handle the influx of data being generated while maintaining accuracy. 

Once eICU data becomes available, this model will be re-evaluated by training on the MIT-BIH database. As described in article \cite{art2}, the model will be transferred to be used on the eICU data being collected. Since the data being generated from the ICU is continuous, after processing 500-1000 samples a comparison of the inference results will occur. Then depending on these results, additional modifications, such as refined preprocessing steps, will need to be made to the model to improve performance.

Even though the other two articles were not used, there were takeaways from each. The most important is that this type of data does not need a complex model to produce satisfactory classification results. Therefore, moving forward with this project, using the MIT-BIH database is the best to train a model on. Then, by using the model transfer method described in \cite{art2} to infer on live eICU data, a binary classifier can be accurately made.

\printbibliography

\end{document}